\documentclass{article}

\usepackage{PRIMEarxiv}

\usepackage[utf8]{inputenc} 
\usepackage[T1]{fontenc}    
\usepackage[colorlinks=true,linkcolor=blue,citecolor=blue,urlcolor=blue]{hyperref}       
\usepackage{url}            
\usepackage{booktabs}       
\usepackage{amsfonts}       
\usepackage{amssymb}        
\usepackage{nicefrac}       
\usepackage{microtype}      
\usepackage{lipsum}
\usepackage{fancyhdr}       
\usepackage{graphicx}       
\usepackage{amsmath}        
\graphicspath{{media/}}     

\title{A Semantic Segmentation Algorithm for Pleural Effusion Based on DBIF-AUNet
\thanks{\textit{\underline{Citation}}: 
\textbf{Tang et al. A Semantic Segmentation Algorithm for Pleural Effusion Based on DBIF-AUNet. 2025.}} 
}

\author{
  Ruixiang Tang$^{1}$,Mingda Zhang$^{1}$,  Jianglong Qin$^{1,2*}$, Yan Song$^{3,4}$, Yi Wu$^{4}$, Wei Wu$^{5}$ \\
  $^{1}$School of Software, Yunnan University, Kunming 650500 \\
  $^{2}$Yunnan Provincial Key Laboratory of Software Engineering, \\
  Yunnan University, Kunming 650500 \\
  $^{3}$Department of Information Technology, Army Medical University, Chongqing 400038 \\
  $^{4}$Division of Digital Medical Education \& Research, Department of Biomedical \\
  Engineering and Imaging Medicine, Army Medical University, Chongqing 400038 \\
  $^{5}$Department of Thoracic Surgery, Southwest Hospital, Chongqing 400038 \\
  \texttt{*Corresponding author: qinjianglong@ynu.edu.cn}
}

\begin{document}
\maketitle

\begin{abstract}
Pleural effusion semantic segmentation can significantly enhance the accuracy and timeliness of clinical diagnosis and treatment by precisely identifying disease severity and lesion areas. Currently, semantic segmentation of pleural effusion CT images faces multiple challenges. These include similar gray levels between effusion and surrounding tissues, blurred edges, and variable morphology.To address these challenges, we propose the Dual-Branch Interactive Fusion Attention model (DBIF-AUNet). This model constructs a densely nested skip-connection network and innovatively refines the Dual-Domain Feature Disentanglement module (DDFD). The DDFD module orthogonally decouples the functions of dual-domain modules to achieve multi-scale feature complementarity and enhance characteristics at different levels. Concurrently, we design a Branch Interaction Attention Fusion module (BIAF) that works synergistically with the DDFD. This module dynamically weights and fuses global, local, and frequency band features, thereby improving segmentation robustness. Furthermore, we implement a nested deep supervision mechanism with hierarchical adaptive hybrid loss to effectively address class imbalance. Through validation on 1,622 pleural effusion CT images from Southwest Hospital, DBIF-AUNet achieved IoU and Dice scores of 80.1\% and 89.0\% respectively. These results outperform state-of-the-art medical image segmentation models U-Net++ and Swin-UNet by 5.7\%/2.7\% and 2.2\%/1.5\% respectively, demonstrating significant optimization in segmentation accuracy for complex pleural effusion CT images.
\end{abstract}


\section{Introduction}

Pleural effusion is a common clinical condition resulting from various etiologies, including heart failure, infection, and malignant tumors, which lead to fluid accumulation in the pleural cavity between the parietal and visceral pleura \cite{kumari2022}. Accurate identification and segmentation of pleural effusion are crucial for assessing its severity and guiding therapeutic interventions. Currently, traditional pleural effusion identification relies primarily on manual interpretation, which suffers from low efficiency, strong subjectivity, and limited sensitivity in detecting small amounts of effusion \cite{grimberg2010}. These limitations are particularly evident in mesothelioma diagnosis, where diagnostic accuracy and sensitivity remain suboptimal \cite{chee2023}.

In recent years, deep learning technology has achieved significant progress in medical image semantic segmentation \cite{han2024}. Methods based on convolutional neural networks (CNNs) have demonstrated strong performance across various medical image segmentation tasks \cite{li2023}. However, pleural effusion presents unique challenges in CT images, as it often displays similar or gradually transitioning gray values with lung parenchyma, diaphragm, chest wall, and other structures. This similarity causes traditional segmentation methods to fail, especially when effusion coexists with atelectasis and pleural thickening, which further reduces edge discrimination. Moreover, small amounts of effusion ($<$300ml) may only appear as blurred posterior costophrenic angles in supine CT, while large amounts can occupy the entire chest cavity and change from crescent to circular shapes. This morphological variability significantly increases the generalization difficulty for segmentation models. Therefore, precise segmentation remains a considerable challenge. To address these limitations, this paper proposes a network based on dual-domain module decoupling and branch interactive attention fusion (Dual-Branch Interactive Fusion Attention U-Net, DBIF-AUNet) for semantic segmentation of pleural effusion. This model leverages the synergistic advantages of dual-domain feature decoupling and multi-branch interactive attention mechanisms to better handle complex pleural effusion CT images.

\section{Related Work}
\label{sec:related}

\subsection{U-shaped Deep Learning Networks}

U-shaped deep learning networks, particularly U-Net and its variants, have been extensively studied and successfully applied in biomedical image segmentation. U-Net was first proposed by Olaf Ronneberger et al. in 2015 \cite{ronneberger2015}, featuring a contracting path (encoder) and a symmetric expanding path (decoder) that form a distinctive U-shaped structure. The contracting path captures contextual information, while the expanding path enables precise localization. Skip connections allow the network to transfer feature maps from the contracting path to the symmetric expanding path, thereby better preserving image details. Building upon U-Net, Attention U-Net introduces attention mechanisms that help the network focus more effectively on important regions in the image, thus improving segmentation results \cite{lei2024}.

U-Net++ further enhances feature reuse and fusion by introducing nested and dense skip connections \cite{zhou2019}. To overcome the limitations of convolutional neural networks in learning global and long-range semantic information interactions, Hu Cao et al. proposed Swin-UNet, which integrates a Transformer-based architecture into the U-Net structure \cite{cao2021}. However, skip connections in traditional U-shaped deep learning networks directly concatenate shallow and deep features, inevitably leading to semantic gaps and detail loss. To address this issue, we propose a cross-level feature aggregation network with densely nested skip connections, which significantly improves multi-scale feature fusion efficiency through cross-level feature transfer via skip connections.

\subsection{Dual-Domain Feature Fusion}

Traditional semantic segmentation methods primarily focus on spatial domain features of images, namely spatial relationships between pixels and local texture information \cite{lin2024}. To address the limitations of single-domain approaches, researchers have begun exploring the use of features from other domains, particularly the frequency domain, to assist semantic segmentation. Frequency domain features in dual-domain modules can provide global structure and texture information of images, effectively complementing spatial domain features and thus improving segmentation robustness \cite{wei2024}. While spatial domain features describe local texture and spatial relationships of images, frequency domain features capture global structure and frequency information \cite{wei2024, zeng2024}. By fusing spatial and frequency domain features, models can simultaneously focus on both local details and global structure of images, thereby improving segmentation accuracy. Dual-domain feature fusion has achieved significant results in multiple semantic segmentation tasks, including remote sensing image segmentation, medical image segmentation, and autonomous driving scene segmentation \cite{lin2024, wei2024}. However, current traditional dual-domain methods only enhance global features through frequency domain channel attention without explicitly decoupling structure and texture information, resulting in feature confusion problems. We orthogonally decouple traditional dual-domain feature fusion modules and design a dual-domain feature disentanglement module to effectively handle complexity and edge blurriness in images while achieving multi-scale feature complementarity.

\subsection{Attention Mechanisms in Medical Image Segmentation}

In deep learning, attention mechanisms enable models to focus on the most important features in input data by learning weight distributions, thereby improving model performance and efficiency \cite{hassanin2024}. These mechanisms have been widely applied in medical image analysis with remarkable success. In lung CT image semantic segmentation, Yang et al. designed AttGGO-Net, which utilizes self-attention and cross-attention mechanisms to improve segmentation accuracy \cite{yang2023}. Chen et al. proposed a robust Attention-Guided and Noise-Resistant framework (AGNR), which integrates attention-guided mechanisms and noise resistance learning to enhance segmentation accuracy \cite{chen2024}.

For multi-scale attention mechanism fusion, SMANet employs superpixel-guided multi-scale attention, dividing medical images into different regions through superpixel segmentation and using multi-scale attention mechanisms to fuse features from these regions, achieving more precise segmentation \cite{shen2025}. To address the problem that traditional attention mechanisms tend to over-focus on globally most salient effective features while suppressing secondary salient features, Zhan et al. proposed FSA-Net, which improves medical image segmentation by releasing global suppression information \cite{zhan2023}. However, current attention mechanisms struggle to balance global structure, local details, and frequency domain features when dealing with complex edges, and static weight allocation lacks dynamic adaptability. We propose a multi-branch interactive attention fusion module that employs three-level feature information processing and gated interactive selection mechanisms, significantly enhancing the ability to delineate complex edges and overcoming the locality limitations of traditional single attention.

\subsection{Deep Supervision}

Deep supervision technology has been widely applied in semantic segmentation by introducing intermediate layer supervision in neural network training to strengthen learning signals and overcome gradient vanishing problems in deep network training \cite{zhao2023, yuan2023}. Due to the complexity of deep networks and their dependence on large amounts of annotated data, the training process often faces problems such as gradient vanishing or convergence difficulties. Deep supervision technology improves model training efficiency and performance by optimizing intermediate layer feature representations and transmitting supervision signals more directly to intermediate layers of the network. In medical image segmentation, deep supervision has significantly improved segmentation accuracy for various pathologies \cite{dzieniszewska2025, liu2025, maqsood2025}, while simultaneously accelerating network convergence speed and improving final model performance by providing additional learning signals \cite{li2022}. However, current deep supervision approaches (such as UNet++) \cite{zhou2019} use identical loss functions at different levels, ignoring semantic differences between levels and struggling with class imbalance problems. We design a nested deep supervision mechanism to address these issues, effectively improving segmentation accuracy for small targets and alleviating gradient vanishing problems.

\section{Method}

\subsection{Nested Skip Connection Structure of DBIF-AUNet}

We propose DBIF-AUNet (Dual-Branch Interactive Fusion Attention U-Net) for handling images with complex and diverse edges. DBIF-AUNet adopts a nested skip connection U-shaped network architecture similar to UNet++ \cite{zhou2019} and ANUNet \cite{li2020}. Figure 1 illustrates the densely nested skip connection structure of DBIF-AUNet.

The improved densely nested skip connection structure consists of five essential parts: the DBIF-AUNet encoder, decoder, DDFD dual-domain feature fusion module, BIAF branch interactive attention fusion module, and nested deep supervision module.

As shown in Figure 1, cross-level skips first aggregate deep features and shallow features with current layer features. Note that this figure only displays the outermost skip connection structure, while the internal skip connection method follows the same pattern as the outermost. Subsequently, the input multi-level features are processed through the DDFD module with functionally decoupled three independent sub-modules to obtain deep semantics, current layer refined features, and shallow details. To avoid information loss caused by feature transfer in the decoder in traditional connection methods, the DDFD module directly transfers nested features to the three input branches of the BIAF module, which then processes feature information from different layers for interactive dynamic fusion.

\begin{figure}[htbp]
  \centering
  \includegraphics[width=0.8\textwidth]{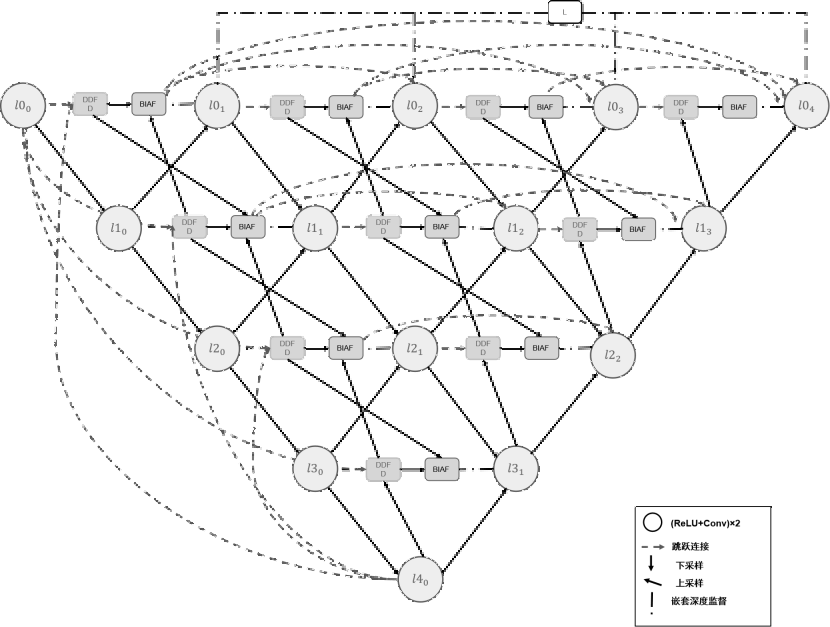}
  \caption{Nested skip connection structure of DBIF-AUNet}
  \label{fig:figure1}
\end{figure}

\subsection{Dual-Domain Feature Disentanglement Module}

To address the structural complexity of pleural effusion images, we select dual-domain modules that can combine spatial and frequency domains to enhance feature information. To solve the confusion limitations of traditional dual-domain fusion, we design a module decoupling enhancement method for multi-level features, performing targeted processing on features at different levels. The dual-domain module is functionally orthogonally decomposed into a dual-domain decoupling module, which strengthens boundary perception through spatial strip attention and combines with frequency domain band attention decoupling. Subsequently, it achieves multi-scale feature complementarity through bidirectional frequency band modulation and spatial-frequency domain alignment.

As shown in Figure 2, the deep, shallow, and current layer features input from skip connections are processed through the DDFD (Dual-Domain Feature Disentanglement) module after feature aggregation and normalization. For the shallow features of layer l0, these consist of shallow convolution structures with 3×3 convolution kernel plus ReLU activation. The three dual-domain module decoupling functional modules in the DDFD module explicitly separate semantic information at different levels. Specifically, the global branch captures the overall morphology of the image, the local branch strengthens edge gradient changes, and the channel branch filters noise and enhances cross-channel texture expression. The processing methods for each branch are as follows:

\begin{figure}[htbp]
  \centering
  \includegraphics[width=0.9\textwidth]{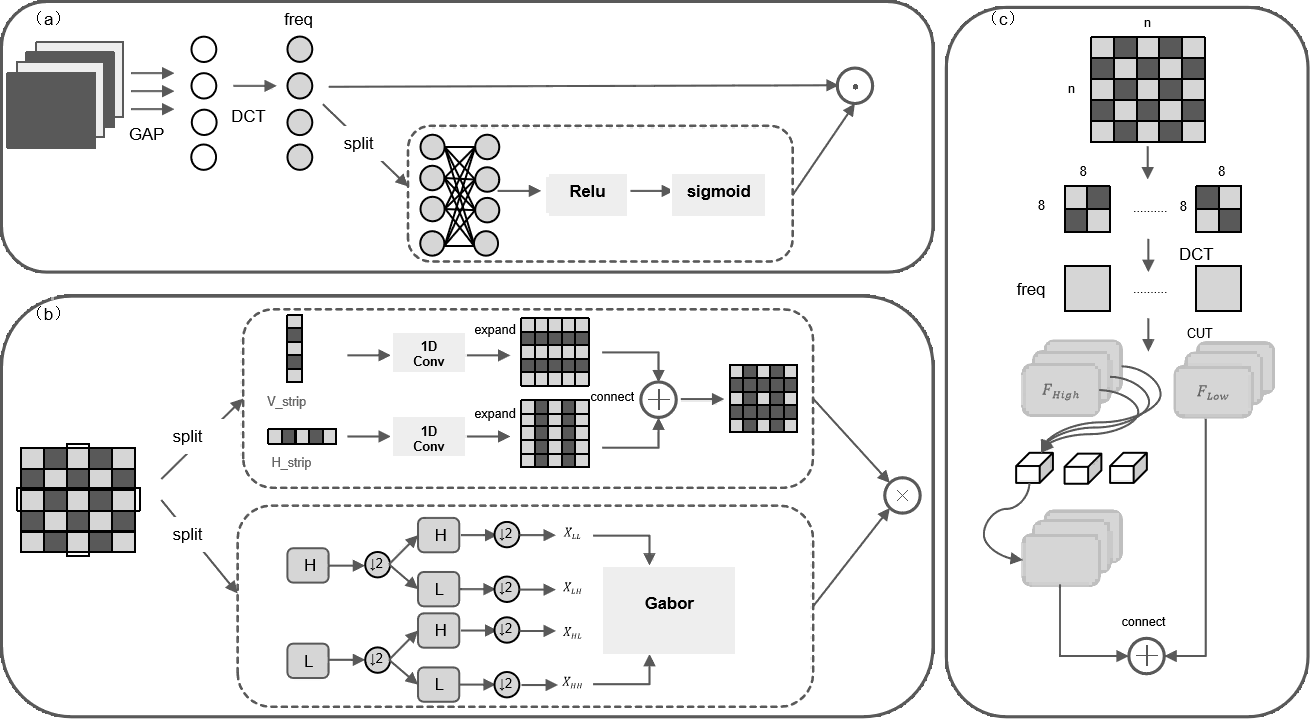}
  \caption{Decoupling functional structure of the dual-domain feature disentanglement module}
  \label{fig:figure2}
\end{figure}

1. Global average pooling (GAP) is used to extract low-frequency global structure of deep features, and after enhancing cross-channel interaction through frequency domain channel attention, global context features with frequency bands are obtained and input to the global branch through bilinear interpolation.

\begin{equation}
\mathbf{c} = \sigma\left( \mathbf{W}_{2} \cdot ReLU\left( \mathbf{W}_{1} \cdot \left| \mathcal{DCT}\left( \frac{1}{H \times W}\sum_{i = 1}^{H}\sum_{j = 1}^{W}\mathbf{X}_{i,j;} \right) \right| + \mathbf{b}_{1} \right) + \mathbf{b}_{2} \right) \odot \mathcal{DCT}\left( \frac{1}{H \times W}\sum_{i = 1}^{H}\sum_{j = 1}^{W}\mathbf{X}_{i,j;} \right)
\end{equation}

where $\sum_{i = 1}^{H}\sum_{j = 1}^{W}X_{i,j;}$ represents calculating spatial sum and output vector independently for each channel, $W_{1}$ and $W_{2}$ are learnable parameters, $\mathcal{DCT}$ is 2D DCT transform, $ReLU$ and $\sigma$ are activation functions, and $\odot$ denotes element-wise multiplication.

2. Boundary perception is strengthened through Strip Pooling, which directly associates shallow detail features with deep semantics and combines spatial position weights with frequency domain energy distribution. After performing 2D discrete wavelet transform (DWT) on input feature $X$ through wavelets, we separate low-frequency approximation component $X_{LL}$ and high-frequency detail component $X_{HH}$. Meanwhile, Gabor filters balance spatial resolution and frequency band resolution, achieving spatial skip attention and spatial-frequency domain balance mechanism to obtain high-frequency edge features. The fusion features are shown in Equation 5, and the results are finally input to the local branch after spatial-frequency domain alignment.

\begin{equation}
\mathbf{y}_{\mathbf{c,i}}^{\mathbf{h}} = \frac{\mathbf{1}}{\mathbf{W}}\sum_{\mathbf{s}}^{\mathbf{W}}\mathbf{X}_{\mathbf{c,i,j}} \in \mathbb{R}^{\mathbf{C \times H \times 1}}
\end{equation}

\begin{equation}
\mathbf{y}_{\mathbf{cj}}^{\mathbf{v}} = \frac{\mathbf{1}}{\mathbf{H}}\sum_{\mathbf{i = 1}}^{\mathbf{H}}\mathbf{X}_{\mathbf{c,i,j}} \in \mathbb{R}^{\mathbf{C \times 1 \times W}}
\end{equation}

\begin{equation}
\mathbf{G}_{\mathbf{\theta,f}}\mathbf{(i,j) =} \exp\left( -\frac{\mathbf{i}^{'2} + \mathbf{\gamma}^{2}\mathbf{j}^{'2}}{\mathbf{2}\mathbf{\sigma}^{2}} \right)\cos(\mathbf{2\pi fi' + \phi})
\end{equation}

where $i' = i\cos\theta + j\sin\theta$ and $j' = -i\sin\theta + j\cos\theta$. Here, orientation parameter $\theta$ and frequency $f$ adaptively adjust spatial and frequency band resolution weights.

\begin{equation}
\hat{\mathbf{X}}_{\mathbf{local}} = Concat(\mathbf{y}^{\mathbf{h}}, \mathbf{y}^{\mathbf{v}}) \otimes \mathbf{G}_{\mathbf{\theta,f}}(\mathbf{X}_{\mathbf{LL}}, \mathbf{X}_{\mathbf{HH}})
\end{equation}

where $\otimes$ represents spatial weighting operation, enhancing collaborative expression of edges and textures.

3. Through 2D DCT transform as shown in Equation 6, we separate low-frequency (structural information) and high-frequency (texture details) of current layer features, along with cross-band texture obtained from frequency band separation. These are then reconstructed and input to the channel branch through DCT-IDCT.

\begin{equation}
\mathbf{F(u,v) = \alpha(u)\alpha(v)} \sum_{\mathbf{i = 0}}^{\mathbf{H - 1}}\sum_{\mathbf{j = 0}}^{\mathbf{W - 1}}\mathbf{X}_{\mathbf{c,i,j}} \cos\left[\frac{\mathbf{\pi(2i + 1)u}}{\mathbf{2H}}\right] \cos\left[\frac{\mathbf{\pi(2j + 1)v}}{\mathbf{2W}}\right]
\end{equation}

where $\alpha(u) = \sqrt{\frac{1}{H}}$ when u=0, or $\sqrt{\frac{2}{H}}$ when u>0.

\subsection{Branch Interactive Attention Fusion Module}

Due to the variable morphology of pleural effusion in CT images, such as diffuse small effusion and encapsulated complex effusion, along with blurred boundaries with surrounding tissues including lung parenchyma and pleura, we design a module capable of multi-level feature fusion and interactive weight adjustment. This module builds upon traditional SE attention modules and combines with feature information processed by the dual-domain decoupling module. The module consists of three components: branch parallel enhancement module, branch attention fusion module, and interactive attention fusion module for weighted feature fusion, which dynamically adjusts weights interactively for different morphologies of complex images. After aligning the feature channels from Section 2.2 and inputting them into the three input branches of this module, the module adopts a three-level architecture of parallel processing, followed by feature fusion, and finally dynamic weighting. By processing features from three scales - global branch for low-frequency structure, local branch for high-frequency edges, and channel branch for frequency band texture - and then performing dynamic interactive fusion, it achieves balanced expression of structure and details in complex pleural effusion CT images, thus enhancing robustness in segmentation environments with variable morphology and blurred boundaries. Figure 3 shows the structure of the BIAF (Branched-Interactive Attention Fusion Module).

\begin{figure}[htbp]
  \centering
  \includegraphics[width=0.6\textwidth]{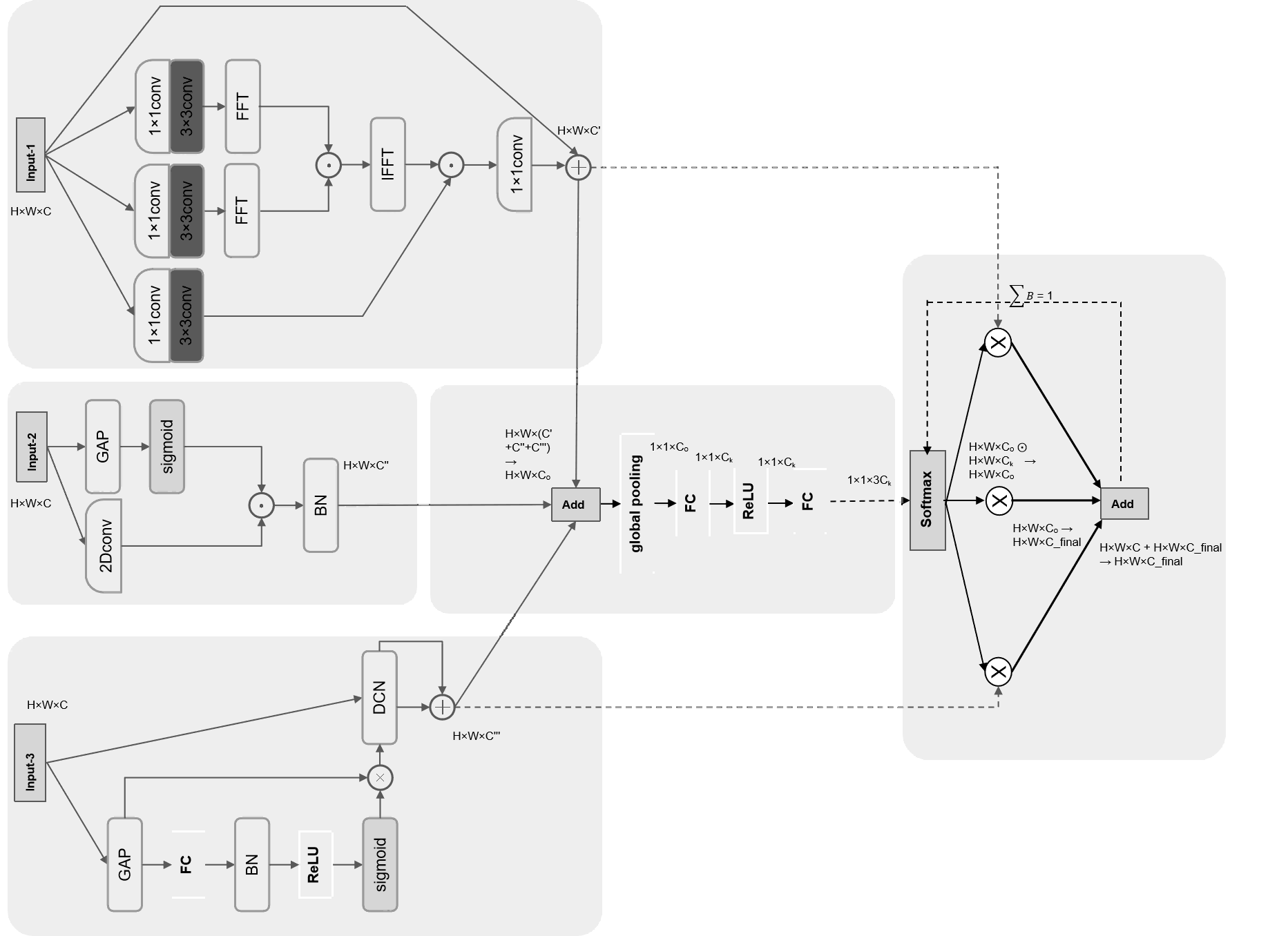}
  \caption{Branch Interactive Attention Fusion Module}
  \label{fig:figure3}
\end{figure}

The branch fusion module further processes the input feature information before fusion by adopting parallel enhancement of frequency domain-guided global modeling strengthening, local deformation adaptive enhancement, and channel-aware noise filtering. The processing procedure is shown in the following formulas:

The global branch uses frequency domain self-attention to transform features to frequency domain through FFT, establishing long-range dependencies and solving the locality limitations of spatial convolution. Meanwhile, cross-band complementarity strengthens the interaction from low-frequency contour information to mid-frequency texture information to high-frequency edge details, avoiding loss of key information:

\begin{equation}
\mathbf{A}_{\mathbf{g}} = \mathcal{CFC}( Softmax(\frac{\mathcal{F}(\mathbf{Q}) \cdot \mathcal{F}(\mathbf{K})^{\mathbf{T}}}{\sqrt{\mathbf{d}_{\mathbf{k}}}}))
\end{equation}

where $\mathcal{CFC}()$ denotes the cross-band correlation function.

The local branch adopts local feature adaptive dynamic convolution groups that adaptively adjust convolution kernel parameters according to input content, enhancing robustness to deformation occlusion. Then it uses frequency band normalization to group and standardize features at different frequencies, solving training instability caused by frequency domain distribution differences:

\begin{equation}
\mathbf{A}_{\mathbf{l}} = \frac{\mathbf{1}}{\mathbf{K}}\sum_{\mathbf{k = 1}}^{\mathbf{K}} Sigmoid(\mathbf{U}_{\mathbf{k}} \cdot GAP(X))(\mathbf{X} * \mathbf{W}_{\mathbf{k}}) \oslash \sqrt{\mathbf{\sigma}_{\mathbf{k}}^{2} + \epsilon}
\end{equation}

where $\sigma_{k}^{2}$ is the variance of the k-th frequency band, $\oslash$ denotes element-wise division, and $\mathbf{U}_{k}$, $\mathbf{W}_{k}$ are convolution kernel parameters.

The channel branch uses deformable convolution to learn spatial offsets, adapting to non-rigid deformations, while adjusting channel soft thresholds through learnable thresholds to filter irrelevant frequency band noise:

\begin{equation}
\mathbf{A}_{\mathbf{c}} = \eta\left( \sum_{\mathbf{n} \in \mathbb{N}} \mathbf{w}_{\mathbf{n}} \cdot X(\mathbf{p}_{\mathbf{0}} + \mathbf{p}_{\mathbf{n}} + \Delta\mathbf{p}_{\mathbf{n}}) \right) \circ \eta\left( \sum_{\mathbf{m} \in \mathcal{M}} \mathbf{v}_{\mathbf{m}} \cdot \mathbf{X}(\mathbf{q}_{\mathbf{0}} + \mathbf{q}_{\mathbf{m}} + \Delta\mathbf{q}_{\mathbf{m}}) \right)
\end{equation}

where $\eta(z) = sign(z) \cdot \max(0, |z| - \tau)$, $p_{n}$ and $q_{m}$ represent offsettable spatial variables, and $\circ$ is the composite function.

Subsequently, dynamic fusion is performed through the branch attention fusion module:

\begin{equation}
\mathbf{Y} = \sigma\left( \mathbf{W}_{\mathbf{f}_{2}}^{\top} \cdot ReLU\left( \mathbf{W}_{\mathbf{f}_{1}}^{\top} \cdot GAP(Z) \right) \right) + Concat(\mathbf{A}_{\mathbf{g}} * \mathbf{X}_{\mathbf{1}}, \mathbf{A}_{\mathbf{l}} * \mathbf{X}_{\mathbf{2}}, \mathbf{A}_{\mathbf{c}} * \mathbf{X}_{\mathbf{3}})
\end{equation}

where $W_{f_{1}}^{\top}$ and $W_{f_{2}}^{\top}$ are fully connected weights.

Finally, three-branch decoupling is performed in the interactive attention fusion module, where weights are normalized through the Softmax function and interactive dynamic weighted output is performed as shown in the following formula:

\begin{equation}
\mathbf{B}_{\mathbf{i}} = \frac{\exp(\mathbf{W}_{\mathbf{b}_{\mathbf{i}}}^{\top} \cdot Y)}{\sum_{\mathbf{j = 1}}^{3} \exp(\mathbf{W}_{\mathbf{b}_{\mathbf{j}}}^{\top} \cdot Y)} \quad \text{for} \quad i = 1,2,3
\end{equation}

where $W_{b_{i}}^{\top}$ represents channel attention weight.

\begin{equation}
\mathbf{O} = \sum_{\mathbf{i = 1}}^{3} \mathbf{B}_{\mathbf{i}} \odot \mathbf{Y}_{\mathbf{i}}
\end{equation}

Finally, interactive gated selection is performed for three-way information, where $B_{i}$ denotes the weight vector.

\subsection{Nested Deep Supervision}

To implement the nested deep supervision structure designed in this paper, we develop a multi-level nested deep supervision hybrid loss. In addition to adding supervision nodes at the end of each upsampling path in the original UNet++ deep supervision mechanism, we also insert deep supervision points in each BIAF module output path, solving the multi-level supervision balance problem through adaptive weights. The calculation process of the loss function for this method proceeds as follows:

First, we calculate a hybrid loss at each supervision point using Dice Loss, Focal Loss, and Binary Cross-Entropy Loss respectively. These losses handle class imbalance and optimize segmentation boundaries, focus on hard samples, and provide stable gradients. The weighted hybrid loss function is shown in Equation 16.

\begin{equation}
\mathbf{L}_{\mathbf{Dice}}^{(k)} = 1 - \frac{2\sum(\mathbf{Y}_{\mathbf{true}} \odot \mathbf{Y}_{\mathbf{pred}}^{(k)}) + \epsilon}{\sum \mathbf{Y}_{\mathbf{true}} + \sum \mathbf{Y}_{\mathbf{pred}}^{(k)} + \epsilon}
\end{equation}

where $Y_{true}$ represents the ground truth label and $Y_{pred}^{(k)}$ represents the prediction.

\begin{equation}
\mathbf{L}_{\mathbf{Focal}}^{(k)} = -\frac{1}{N}\sum_{i = 1}^{N}[\alpha(1 - \mathbf{Y}_{\mathbf{pred,i}}^{(k)})^{\gamma} \mathbf{Y}_{\mathbf{true,i}} \log(\mathbf{Y}_{\mathbf{pred,i}}^{(k)}) + (1 - \alpha)(\mathbf{Y}_{\mathbf{pred,i}}^{(k)})^{\gamma}(1 - \mathbf{Y}_{\mathbf{true,i}}) \log(1 - \mathbf{Y}_{\mathbf{pred,i}}^{(k)})]
\end{equation}

\begin{equation}
\mathcal{L}_{\mathbf{BCE}}^{(k)} = -\frac{1}{N}\sum_{i = 1}^{N}[\mathbf{Y}_{\mathbf{true,i}}^{(k)} \log(\mathbf{Y}_{\mathbf{pred,i}}^{(k)}) + (1 - \mathbf{Y}_{\mathbf{true,i}}^{(k)}) \log(1 - \mathbf{Y}_{\mathbf{pred,i}}^{(k)})]
\end{equation}

\begin{equation}
\mathcal{L}_{\mathbf{hybrid}}^{(k)} = \lambda_{\mathbf{Dice}} \mathcal{L}_{\mathbf{Dice}}^{(k)} + \lambda_{\mathbf{Focal}} \mathcal{L}_{\mathbf{Focal}}^{(k)} + \lambda_{\mathbf{BCE}} \mathcal{L}_{\mathbf{BCE}}^{(k)}
\end{equation}

where hyperparameter weights $\lambda_{Dice}$, $\lambda_{Focal}$, and $\lambda_{BCE}$ are set to 0.4, 0.3, and 0.3 respectively.

The supervision point function is defined as:

\begin{equation}
\mathbf{H(X)} = \mathbf{Conv}_{1 \times 1} \circ ReLU \circ \mathbf{Conv}_{3 \times 3} \circ Sigmoid(X)
\end{equation}

Upsampling endpoint supervision is calculated as:

\begin{equation}
\mathcal{L}_{\mathbf{U}_{j}} = \mathcal{L}_{\mathbf{hybrid}}(\mathcal{H}(\mathbf{I}_{0j}), \mathbf{Y}_{\mathbf{true}})
\end{equation}

BIAF output path loss is computed as:

\begin{equation}
\mathcal{L}_{\mathbf{B}_{j}} = \mathcal{L}_{\mathbf{hybrid}}(\mathbf{Deconv}_{2j}(\mathcal{H}(\mathbf{S}_{j})), \mathbf{Y}_{\mathbf{true}})
\end{equation}

Since different supervision points have varying importance, we assign different weights according to their layer depth, with weight factors as shown in Equation 21:

\begin{equation}
\omega_{r}^{(j)} = \frac{Area(\mathbf{X}_{0,j})}{\sum_{m = 1}^{4} Area(\mathbf{X}_{0,m})} = \frac{4^{j - 1}}{\sum_{m = 1}^{4} 4^{m - 1}}
\end{equation}

\begin{equation}
\mathbf{w}_{\mathbf{S}_{j}} = Softmax(1.5 \cdot \omega_{r}^{(j)} \cdot 0.7)
\end{equation}

\begin{equation}
\mathbf{w}_{\mathbf{U}_{j}} = Softmax(1.5 \cdot \omega_{r}^{(j)} \cdot 0.9)
\end{equation}

The final weighted multi-level supervision hybrid loss function is shown in Equation 24:

\begin{equation}
\mathcal{L}_{\mathbf{Total}} = \sum_{j = 1}^{4} \mathbf{w}_{\mathbf{U}_{j}} \mathcal{L}_{\mathbf{U}_{j}} + \sum_{j = 1}^{4} \mathbf{w}_{\mathbf{S}_{j}} \mathcal{L}_{\mathbf{B}_{j}}
\end{equation}

\section{Experiments}

\subsection{Dataset}

Due to the lack of high-quality public 2D pleural effusion CT image datasets, the dataset used in this paper comes from 9 expert-annotated patient imaging data containing various thoracic diseases authorized by Southwest Hospital. The dataset includes 1,622 DICOM format CT files and supporting Amira annotation files, with ethics approval number (B)KY2023062. All annotations were completed by thoracic surgeons and verified for consistency with a kappa value of 0.87, and sensitive privacy information was removed from the data. The dataset processing workflow is as follows:

\textbf{Standardization and Conversion:} We unified DICOM files with window width and level settings using lung window parameters of 1500/-600 HU, and standardized voxel spacing to 1.0×1.0×1.0 mm³ through bilinear interpolation. All files were converted to 512×512 pixel 24-bit PNG format CT images, totaling 1,622 images. Simultaneously, we unified Amira annotation file effusion color to RGB values of 255,0,0, then normalized and converted them to 2D PNG format multi-class masks.

\textbf{Binary Extraction:} Based on HSV threshold settings with H$\in$[0,10], S$\in$[200,255], and V$\in$[200,255], we extracted pleural effusion regions from images originally containing various thoracic diseases, generating single-channel binary masks with 0 for background and 255 for foreground.

\textbf{Data Pairing and Normalization:} We strictly aligned CT and mask filenames, linearly normalized CT gray values to the range [0,1], and binarized masks to values of 0 and 1.

\textbf{Stratified Dataset Division:} Using stratified sampling by patient ID to avoid cross-patient data leakage, we divided the dataset into a training set of 1,298 images (80

\textbf{Quality Control:} We randomly sampled 5

The processed dataset images are shown in Figure 4:

\begin{figure}[htbp]
  \centering
  \includegraphics[width=\textwidth]{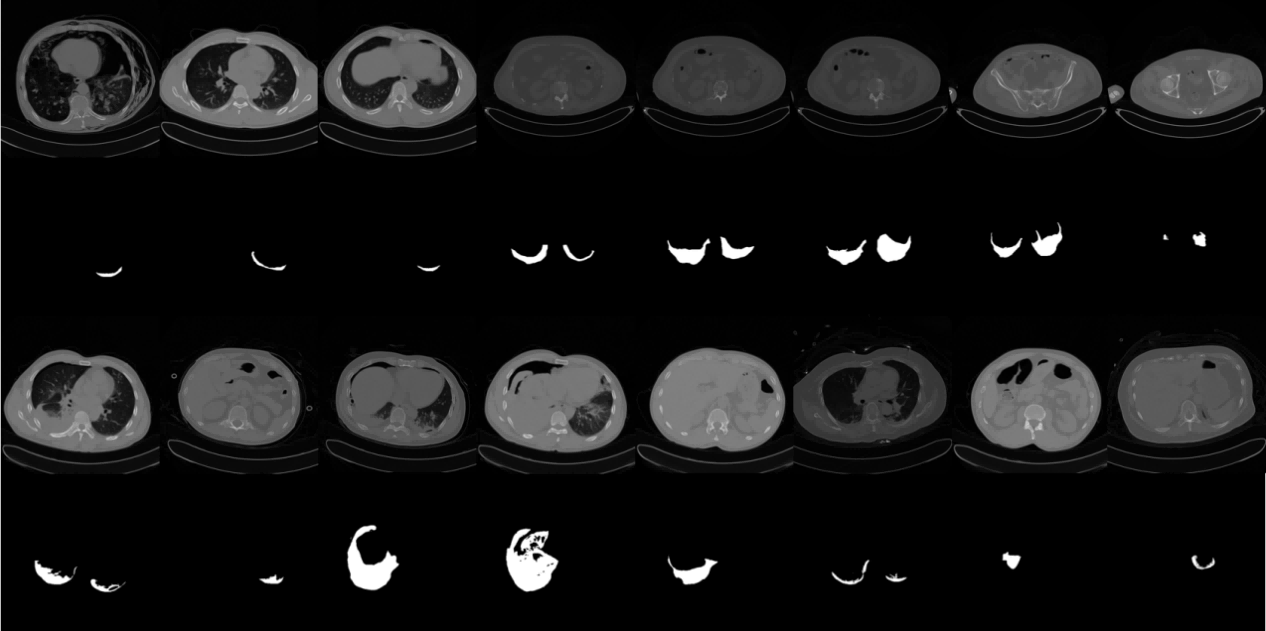}
  \caption{The constructed pleural effusion CT image dataset}
  \label{fig:figure4}
\end{figure}

\subsection{Experimental Parameters}

Our proposed DBIF-AUNet is implemented using the deep learning framework PyTorch 1.7.0 and Python 3.7.10, and trained using Intel® Xeon® E5-2686 v4 CPU and 2×NVIDIA GeForce RTX 4090 GPU with 24GB memory.

The experiment uniformly adjusts image resolution to 512×512 pixels and uses the SGD optimizer to optimize the training model. The batch size is set to 8, 16, or 24 according to different dataset sizes. The initial learning rate is set to 0.001, and we use a cosine annealing learning rate with restart dynamic update strategy to optimize the training process. The restart step size is set to 20, gamma is set to 0.5, and the training runs for 200 epochs.

\subsection{Comparative Experiments}

To verify the performance of our optimized model and the quality of the dataset from multiple perspectives, we train other deep learning networks on the constructed dataset and calculate the segmentation results and evaluation metrics of each network. We select U-Net \cite{ronneberger2015}, UNet++ \cite{zhou2019}, ANUNet \cite{li2020}, and Swin-UNet \cite{cao2021} for comparative experiments. In this paper, we choose Dice Similarity Coefficient (DSC) and Intersection over Union (IoU) as core quantitative evaluation metrics, along with commonly used metrics in semantic segmentation models including Accuracy, Precision, Recall, and Specificity. We also visualize the test set prediction results for comparison.

The calculation formulas for IoU and Dice are shown in Equations 25 and 26 respectively:

\begin{equation}
\mathbf{IoU} = \frac{|A \cap B|}{|A \cup B|}
\end{equation}

\begin{equation}
\mathbf{Dice} = \frac{2 \times |A \cap B|}{|A| + |B|}
\end{equation}

where $A \cap B$ represents the intersection area between prediction and ground truth, and $A \cup B$ represents the union area between prediction and ground truth, calculated as $A+B-A \cap B$. IoU focuses more on segmentation accuracy, while Dice focuses more on addressing class imbalance.

We selected currently excellent medical image segmentation networks in the medical deep learning field for horizontal comparison:

\begin{table}[h]
\centering
\caption{Comparative experimental results}
\label{tab:comparison}
\begin{tabular}{lccccccc}
\toprule
Model & Year & IoU(\%) & Dice(\%) & ACC & Pre & Re & Sp \\
\midrule
U-Net & 2015 & 71.1 & 83.1 & 89.1 & 83.2 & 83.0 & 89.4 \\
UNet++ & 2018 & 74.4 & 85.3 & 90.4 & 85.2 & 85.4 & 90.9 \\
ANU-Net & 2020 & 75.1 & 85.8 & 91.2 & 85.8 & 85.8 & 91.3 \\
Swin-UNet & 2021 & 77.9 & 87.6 & 92.7 & 87.7 & 87.6 & 92.1 \\
DBIF-AUNet (Ours) & - & 80.1 & 89.0 & 94.4 & 89.1 & 89.0 & 94.7 \\
\bottomrule
\end{tabular}
\end{table}

From the data in the table, it can be observed that this dataset achieves good segmentation results on current classical models. Compared to other current medical deep learning models, DBIF-AUNet comprehensively leads in pleural effusion CT image segmentation accuracy with IoU of 80.1

Figure 5 shows the visualization comparison of segmentation results between DBIF-AUNet and the comparison models mentioned above. It can be seen that when facing complex and diverse pleural effusion CT images with unclear edge decomposition for semantic segmentation, our proposed model can more accurately segment complex images and locate boundaries precisely. In contrast, U-Net, UNet++, ANUNet, and Swin-UNet commonly suffer from incorrect segmentation and boundary missing segmentation problems. In summary, the DBIF-AUNet proposed in this paper demonstrates superior segmentation capability on pleural effusion CT images. It can more finely sense changes in image gray values to accurately locate edge boundary segmentation, and shows better accuracy in multi-morphology image recognition without incorrect segmentation for variable effusion types. These results demonstrate its advantages in the field of multi-morphology complex image segmentation with blurred edge boundaries.

\begin{figure}[htbp]
  \centering
  \includegraphics[width=\textwidth]{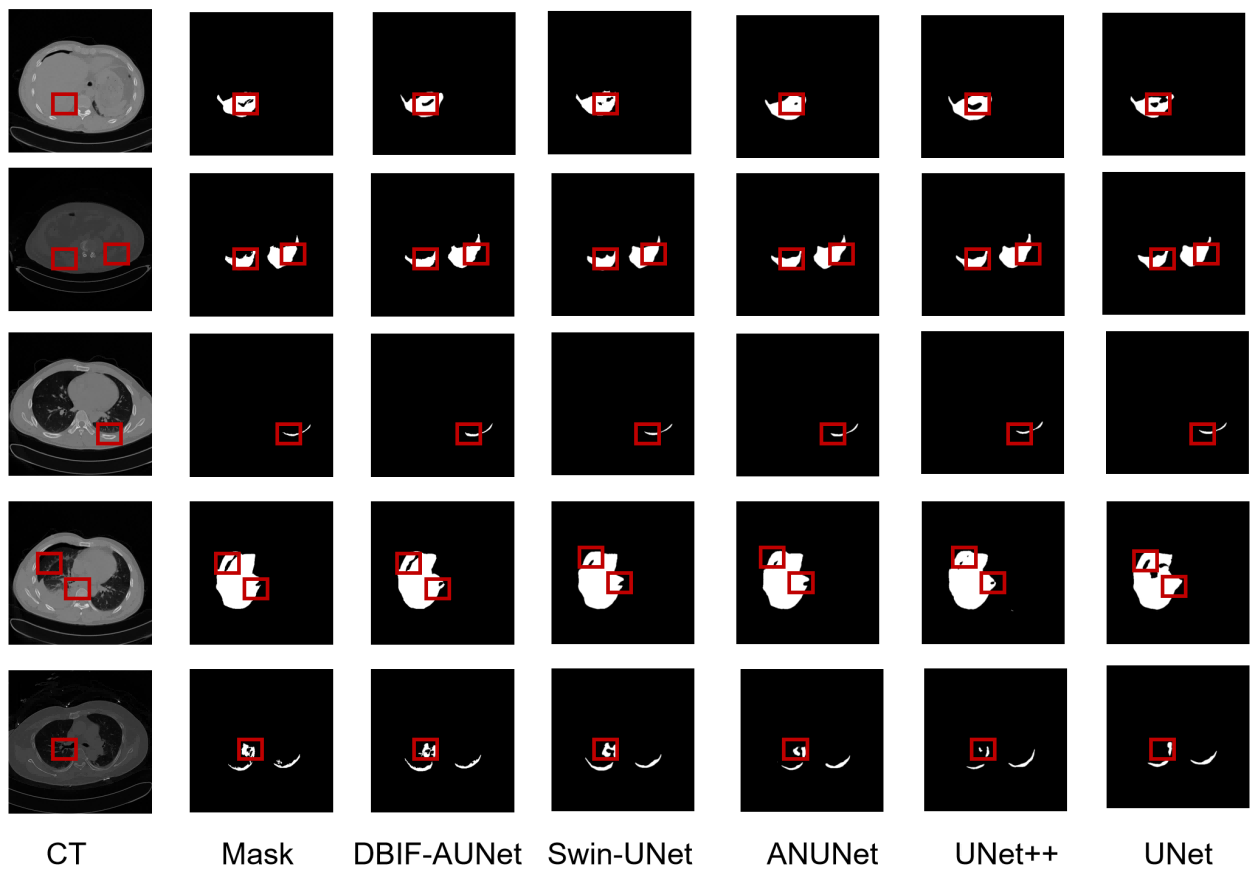}
  \caption{Visual comparison of predictions between DBIF-AUNet and other models}
  \label{fig:figure5}
\end{figure}

\subsection{Ablation Study}

To verify the effectiveness of the proposed method with strict variable control, we continue to use the constructed pleural effusion CT image dataset for comparative experiments.

Table 2 shows the results of the ablation study. We conduct ablation experiments by removing or replacing modules, with specific network structure designs as follows:

\textbf{DBIF-AUNet:} The complete dual-domain decoupling multi-level interactive attention model proposed in this paper.

\textbf{w/o DDFD+BIAF:} This configuration removes the dual-domain decoupling module and multi-level interactive attention fusion module, and deletes the densely nested skip connections designed based on these two modules. It only retains the main network structure and deep supervision module, keeping only the added deep supervision heads after removal.

\textbf{w/o DS:} This variant removes the deep supervision mechanism proposed in this paper, retaining other network architecture and the basic deep supervision mechanism of UNet++.

From the information presented in Figure 6 and Table 2, it can be clearly seen that all modules proposed in this paper contribute to improving segmentation accuracy in the network.

\begin{table}[h]
\centering
\caption{Ablation study results}
\label{tab:ablation}
\begin{tabular}{lcccccc}
\toprule
Module & IoU(\%) & Dice(\%) & ACC & Pre & Re & Sp \\
\midrule
w/o DDFD+BIAF & 74.6 & 85.4 & 90.6 & 85.4 & 85.4 & 91.1 \\
w/o DS & 79.5 & 88.6 & 94.2 & 88.7 & 88.5 & 94.0 \\
DBIF-AUNet (Ours) & 80.1 & 89.0 & 94.4 & 89.1 & 89.0 & 94.7 \\
\bottomrule
\end{tabular}
\end{table}

Figure 6 shows the visual comparison of model results after ablation experiments:

\begin{figure}[htbp]
  \centering
  \includegraphics[width=0.8\textwidth]{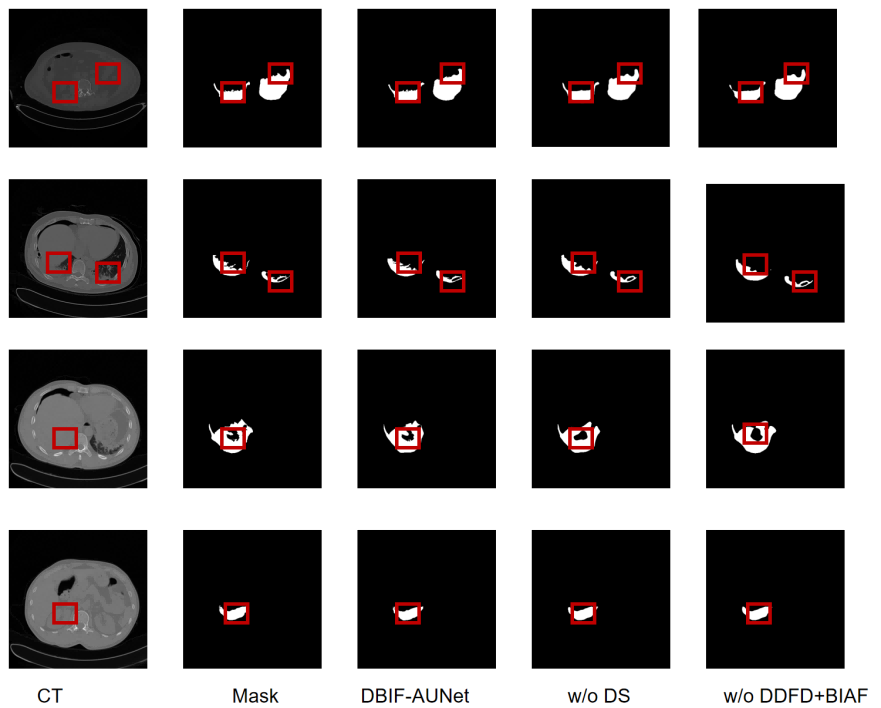}
  \caption{Comparison of ablation experiment results}
  \label{fig:figure6}
\end{figure}

\section{Conclusion and Discussion}

The DBIF-AUNet proposed in this paper effectively addresses the challenges of diverse effusion types and blurred edges in pleural effusion segmentation through dual-domain decoupling and multi-level interactive attention fusion in a densely nested skip connection network. The model demonstrates excellent performance in edge discrimination and morphological adaptability, particularly showing robust segmentation capability for small amounts of effusion with blurred costophrenic angle features and large-volume circular-like lesions. The nested deep supervision mechanism further improves training stability, accelerates model convergence, and enhances accuracy. When compared with current excellent models in medical image segmentation, our model achieves improvements of at least 2.2


\end{document}